\documentclass[10pt,twocolumn,letterpaper]{article}

\usepackage[pagenumbers]{cvpr} 

\usepackage[dvipsnames]{xcolor}

\definecolor{cvprblue}{rgb}{0.21,0.49,0.74}
\usepackage[pagebackref,breaklinks,colorlinks,citecolor=cvprblue]{hyperref}
\usepackage{cuted}
\usepackage{graphicx}
\usepackage{subcaption}
\usepackage{array}
\usepackage{multirow}
\usepackage{comment}
\usepackage{threeparttable}
\usepackage{appendix}
\usepackage[accsupp]{axessibility}

\def\paperID{12} 
\def\confName{CVPR}
\def\confYear{2024}

\title{Unveiling the Ambiguity in Neural Inverse Rendering: \\A Parameter Compensation Analysis}

\author{
    Georgios Kouros$^1$
    \quad
    Minye Wu$^1$
    \quad
    Sushruth Nagesh$^2$ 
    \quad
    Xianling Zhang$^2$ 
    \quad
    Tinne Tuytelaars$^1$\\
    $^1$ KU Leuven \qquad $^2$ Ford Motor Company\\
    {\tt\small $^1$\{georgios.kouros, minye.wu, tinne.tuytelaars\}@esat.kuleuven.be}\\
    {\tt\small $^2$\{snagesh1, xzhan258\}@ford.com}
}

\begin{document}
\maketitle

\begin{strip}\centering
\vspace{-15mm}
\includegraphics[width=\textwidth]{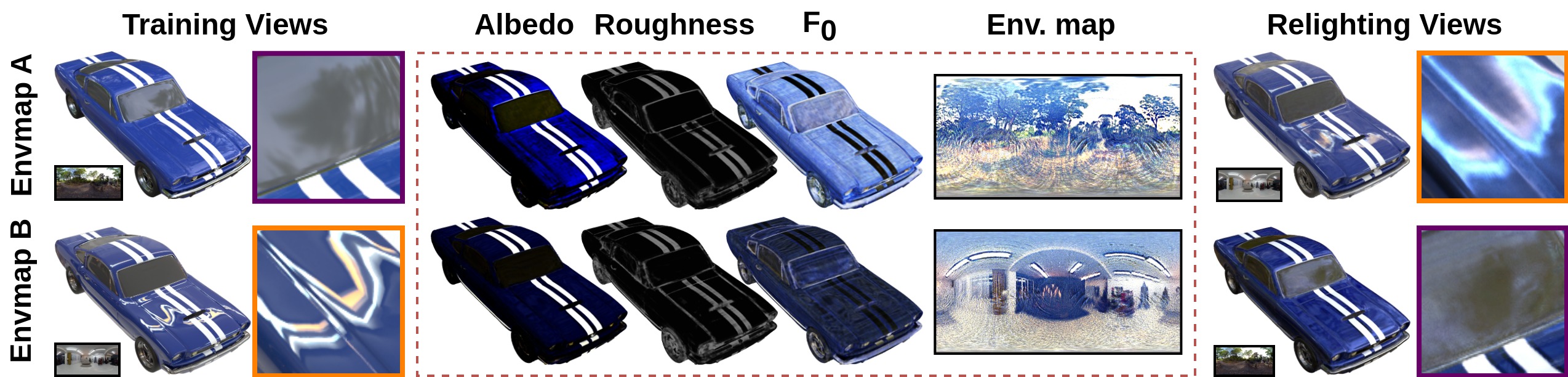}
\vspace{-0.6cm}
\captionof{figure}{
Demonstration of the inherent ambiguity in the task of neural inverse rendering. Training an inverse rendering algorithm like NMF \cite{mai2023nmf} on the same object under different illuminations yields varying material properties that subsequently impact relighting tasks. 
}
\label{fig:teaser}
\end{strip}

\begin{abstract}

Inverse rendering aims to reconstruct the scene properties of objects solely from multiview images. However, it is an ill-posed problem prone to producing ambiguous estimations deviating from physically accurate representations. In this paper, we utilize Neural Microfacet Fields (NMF), a state-of-the-art neural inverse rendering method to illustrate the inherent ambiguity. We propose an evaluation framework to assess the degree of compensation or interaction between the estimated scene properties, aiming to explore the mechanisms behind this ill-posed problem and potential mitigation strategies. Specifically, we introduce artificial perturbations to one scene property and examine how adjusting another property can compensate for these perturbations. To facilitate such experiments, we introduce a disentangled NMF where material properties are independent. The experimental findings underscore the intrinsic ambiguity present in neural inverse rendering and highlight the importance of providing additional guidance through geometry, material, and illumination priors.

\end{abstract}
\section{Introduction}
\label{sec:intro}








Inverse rendering goes beyond simple reflection modeling and involves deducing scene properties from given images, including geometry, lighting, and materials. 
This long-standing challenge finds applications in various fields, such as scene understanding, image manipulation, AR/VR, autonomous driving, etc.
Unfortunately, inferring scene properties solely from multi-view images is an inherently ill-posed inverse problem. 
This is due to the intricate integral relationships defined by the rendering equation, which governs the behavior of light transport and connects geometry, materials, and illumination.
This inherent complexity gives rise to inherent ambiguities in the process.
%

To investigate the inherent ambiguity in the task of neural inverse rendering, we employ Neural Microfacet Fields (NMF) \cite{mai2023nmf}, a state-of-the-art inverse rendering method for efficiently recovering high-fidelity scene geometry, materials, and lighting.
However, similar to most inverse rendering methods the estimated material properties are not always correct when using NMF despite producing plausible reconstructions. NMF, for example, 
may estimate different material properties under different illumination settings, as shown in \cref{fig:teaser}

To understand the correlation between scene properties and improve their estimation, we perform controlled experiments to discover underlying interactions and compensation mechanisms between them. For our experiments, we employ complex scenes of glossy objects from the shiny-blender dataset \cite{verbin2022refnerf} and demonstrate the inherent ambiguity in inverse rendering through inconsistencies among scene properties for different illuminations. To understand the reason behind these inconsistencies, we design an experimental methodology based on which, we perturb one scene property and then fine-tune another to observe how the latter adapts and how well it can compensate to return the model closer to baseline reconstruction performance. We also adapt the architecture to disentangle the material properties and their input features to enable independent fine-tuning. This allows us to analyze the strengths and weaknesses of the complementary relationships between parameters, which in turn elucidates the extent of ambiguity during optimization.
Our contributions can be summarized as follows:
\begin{itemize}
    \item We demonstrate the inherent ambiguity of inverse rendering using NMF, by presenting the inconsistencies in estimated scene properties under different illuminations.
    \item Our key contribution is a framework designed to assess inverse rendering ambiguity, which systematically facilitates the analysis of compensatory dynamics among scene properties.
    \item Leveraging the established framework, our research conducts controlled experiments to identify interactions and compensation mechanisms between scene properties.
\end{itemize}

\section{Related Work}
\label{sec:related}

\subsection*{Neural Radiance Fields}
Neural Radiance Fields (NeRF) \cite{mildenhall2020nerf} have emerged as a groundbreaking method in the field of computer graphics and vision, providing a novel approach to synthesizing photorealistic images from sets of posed 2D images using implicit neural representations.
Although, highly photorealistic and compact, NeRF is inefficient for training and rendering. Subsequent methods adopted hybrid implicit-explicit representations such as voxel grids \cite{SunSC22,sun2022improved,liu2020neural,yu_and_fridovichkeil2021plenoxels} or even more compact tensor decompositions \cite{Chen2022tensorf} or hash-encodings \cite{mueller2022instant} that significantly speed up training and rendering without sacrificing reconstruction quality. Nevertheless, these methods struggle with glossy objects due to the ambiguity in the geometry caused by multi-view inconsistencies. Ref-NeRF\cite{verbin2022refnerf} addressed this limitation by reparameterizing NeRF's view-dependent appearance to better handle reflections, without, however, any relighting capabilities. The lighting-aware NeRF object model trained in LANe \cite{krishnan2023lane} can be composed into scenes with different lighting, but it does not handle glossy material well.

\subsection*{Inverse Rendering}
Inverse rendering is a complex task that aims to reconstruct the properties of a scene, such as geometry, materials, and lighting from images. 
Recent inverse rendering methods have incorporated differentiable volume rendering with implicit neural representations, inspired by NeRF \cite{mildenhall2020nerf}, to learn representations of the scene properties including geometry and materials. Early such works have made strong assumptions or require additional priors, such as a priori known illumination \cite{srinivsan2021nerv,bi2020deepreflectancevolumes} or geometry \cite{zhang2021nerfactor} or  \cite{bi2020deepreflectancevolumes,gao2020deferred} varying illuminations. Such limitations are addressed in  \cite{zhang2021physg,boss2021nerd,boss2021neuralpil,liang2023envidr,fan2023factoredneus,Jin2023TensoIR} by jointly estimating geometry, materials (e.g. BRDF), and lighting. One common issue with inverse rendering methods involves their limited or inefficient ability to handle interreflections, usually through visibility fields \cite{srinivsan2021nerv,zhang2021nerfactor} or radiance transfer fields \cite{guo2020objectcentric, Max1995opticalmodels}. 
NMF \cite{mai2023nmf} 
handles interreflections effectively by casting additional rays from micro-surfaces through the scene using efficient Monte-Carlo sampling with multi-bounce ray tracing.
However, similar to most inverse rendering methods, it is susceptible to ambiguity and estimates plausible yet not consistent material properties under different lighting conditions. 

Using images with varying illuminations has been shown to reduce ambiguity \cite{boss2021nerd,boss2021neuralpil,engelhardt2023-shinobi,yang2022psnerf}. In this work, however, we focus on the constant-illumination setting and what priors can be applied to alleviate the ambiguity of the problem. Priors have been previously used to reduce the ambiguity in inverse rendering and thus produce more consistent and plausible geometry, material properties, and illumination. The different kinds of priors can be categorized depending on which scene property they target, into geometry priors, material priors, and illumination priors.  Geometry priors can be as simple as smoothing priors \cite{guo2022nerfren}, object-category priors \cite{rematasICML21}, or a priori known geometry \cite{zhang2021nerfactor}. Material priors usually involve pretraining of implicit neural BRDFs \cite{boss2021neuralpil,liang2023envidr} on BRDFs of known materials to constrain the search space to plausible materials. Illumination priors, on the other hand, can similarly involve smoothing priors to eliminate high-frequency noise \cite{zhang2021nerfactor,lyu2022neuraltransferfields}, or pretraining to lead the optimization into plausible illuminations \cite{lyu2023diffusionprior,gardner2022rotationequivariant}. 

\section{Methodology}
\label{sec:methodology}

\begin{figure*}[t]
\centering
\includegraphics[width=\textwidth]{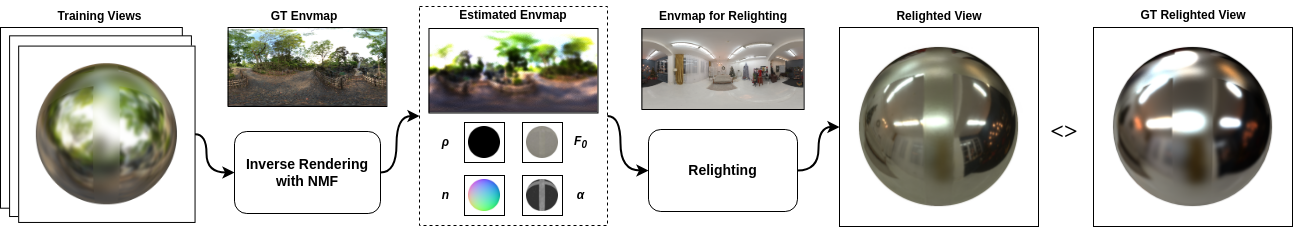}
\vspace{-0.6cm}
\caption{An example of suboptimal estimation of material properties and illumination. The model underestimates the roughness of the ball and overestimates the smoothness of the environment map, which results in sharper reflections when relighting the scene.
}
\label{fig:relighting-issues}
\end{figure*}

\subsection*{Neural Microfacet Fields}
NMF \cite{mai2023nmf} is a neural inverse rendering method that is trained on a collection of posed RGB images and estimates a scene representation composed of volume density $\sigma$, surface normals $\hat{n}$, material properties (BRDFs) including albedo $\rho$, roughness $\alpha$, and reflectance at the normal incidence angle $\mathit{F}_0$. The method also estimates the scene illumination, under a far-field assumption, represented as an environment map. NMF combines volume and surface rendering by treating each 3D point in space as a local micro-facet. This is modeled as a microfacet field represented by a density and a local micro-surface per 3D point. Light is accumulated along rays according to conventional volume rendering, but the outgoing light of every point along a ray is sampled based on its estimated micro-surface.\\

\noindent\textbf{Overall structure.} The geometry and appearance information of the scene is parameterized using TensoRF \cite{Chen2022tensorf}, a low-rank vector-matrix (VM) decomposition of a 4D tensor into multiple compact low-rank tensor components, that ensures storage and training/rendering time efficiency. Specifically, NMF employs one TensoRF model for the density field, noted as $\mathcal{G}_\sigma$, and one to represent the spatially varying appearance features of the scene, noted as $\mathcal{G}_c$. Therefore, the Neural Microfacet Model is designed to decode these features into RGB values and use volume rendering to accumulate pixel colors. \\


\noindent\textbf{Neural Microfacet Model.} The Neural Microfacet Model represents materials through a spatially varying BRDF model corresponding to the sum of the diffuse and specular components of the scene, expressed as

\begin{equation}
f(\hat{\boldsymbol{\omega}}_o, \hat{\boldsymbol{\omega}}_i) = \frac{\rho}{\pi} (1 - Fr(\hat{\boldsymbol{h}})) + F_r(\hat{\boldsymbol{{h}}}) f_s(\hat{\boldsymbol{\omega}}_o, \hat{\boldsymbol{\omega}}_i),
\end{equation}
where $\rho$ is the RGB albedo, $F_r$ is the fresnel term, $\hat{\boldsymbol{h}}$ is the half vector, $f_s$ is the specular component, and $\hat{\boldsymbol{\omega}}_o$,$\hat{\boldsymbol{\omega}}_i$ are the outgoing and incoming light directions. 
NMF uses the Schnell approximation of the Fresnel term, calculated as
\begin{equation}
F_r(\hat{\boldsymbol{{h}}}) = F_0(\boldsymbol{p}) 
+ (1 - F_0(\boldsymbol{p}))
(1 - \hat{\boldsymbol{h}} \cdot \hat{\boldsymbol{\omega}}_o)^5,
\end{equation}
where $F_0(\boldsymbol{p})$ is the specular reflection at the normal incidence for a 3D point $\boldsymbol{p}$.

The specular BRDF, based on the Cook-Torrance model, is expressed as
\begin{equation}
f_s(\hat{\boldsymbol{{\omega}}}_o, \hat{\boldsymbol{{\omega}}}_o) = \frac{
D(\hat{\boldsymbol{h}}; \alpha, \hat{\boldsymbol{n}}, \hat{\boldsymbol{{\omega}}}_o)
G_1(\hat{\boldsymbol{{\omega}}}_o, \hat{\boldsymbol{h}})
g(\hat{\boldsymbol{{\omega}}}_o, \hat{\boldsymbol{{\omega}}}_i)
}{
4 (\hat{\boldsymbol{{n}}} \cdot \hat{\boldsymbol{{\omega}}}_o) 
(\hat{\boldsymbol{{n}}} \cdot \hat{\boldsymbol{{\omega}}}_i)
},
\end{equation}
where $D$ is the Trowbridge-Reitz distribution, an approximation of the material's roughness, $G_1$ is the Smith shadow masking function, and $g$ represents material properties that are not modeled explicitly, but rather implicitly via an MLP.

Albedo $\rho$, roughness $\alpha$, and normal incidence specular reflectance $F_0$ are modeled using MLPs. These MLPs are conditioned on the positional encoding of 3D positions and the appearance features at those positions, and their outputs are passed through sigmoid nonlinearities. The surface normals, on the other hand, are estimated as the negative normalized gradient of the density and are regularized, similar to \cite{verbin2022refnerf} to penalize back-facing normals and prevent semi-transparent surfaces. For more details about the architecture and rendering pipeline of NMF, we refer the reader to the original paper \cite{mai2023nmf}. 

\subsection*{The Ambiguity of Scene Properties}
As already discussed in several existing works \cite{mai2023nmf,lyu2023diffusionprior,zhang2021nerfactor,boss2021nerd,boss2021neuralpil} inverse rendering is a highly ill-posed problem entailing a lot of ambiguity in the estimation of a scene's properties. Consequently, most works \cite{CVPR_2022_Zhang, NEURIPS2023_LightSim} focus on estimating scene properties that produce a plausible reconstruction of the scene, even with suboptimal and physically unrealistic material properties. Although the actual reconstruction may be of high quality, if the geometry and material properties of the scene are not estimated correctly, any downstream relighting task will diverge from the expected behaviour. 

Ambiguity intensifies when dealing with a scene under constant lighting conditions rather than varying. Varying lighting conditions allow a model \cite{krishnan2023lane} to better disambiguate the geometry, material properties, and illumination of the scene. In the absence of varying lighting conditions, however, the estimated material properties can vary depending on how the scene is illuminated. 
%
%
This raises the question of how various scene properties interact or compensate for each other and the extent of their significance. For example, as shown in Figure~\ref{fig:relighting-issues}, the model may underestimate the roughness of the object and produce sharper reflections during relighting tasks. As a result, first, we aim to demonstrate the ambiguity of inverse rendering using NMF as our testbench. Second, we aim to identify potential interactions between the scene properties and how the suboptimal estimation of one can be compensated by another.

To demonstrate the ambiguity and ill-posedness of inverse rendering we investigate how (in)consistent NMF is
by generating the same objects under the same poses but different lighting conditions. Then training NMF models on each of those lighting conditions allows us to observe the variance in the estimated material properties.

Given the established ambiguity in inverse rendering, we aim to investigate how the ambiguity of one scene property affects the other properties and how one compensates for the other so that the model achieves good reconstruction even under suboptimal geometry, material, or illumination estimates. To this end, we attempt to isolate the interaction between two scene properties at a time, by first manipulating one scene property, freezing all other properties but one, and then fine-tuning the remaining unfrozen property. For this experiment, we select the view-consistent material properties of the model, namely albedo $\rho$, roughness $\alpha$, and normal incidence reflectance $F_0$, as well as the density $\sigma$ and the estimated environment map.\\

\begin{figure}[t]
    \centering
    \includegraphics[width=\linewidth]{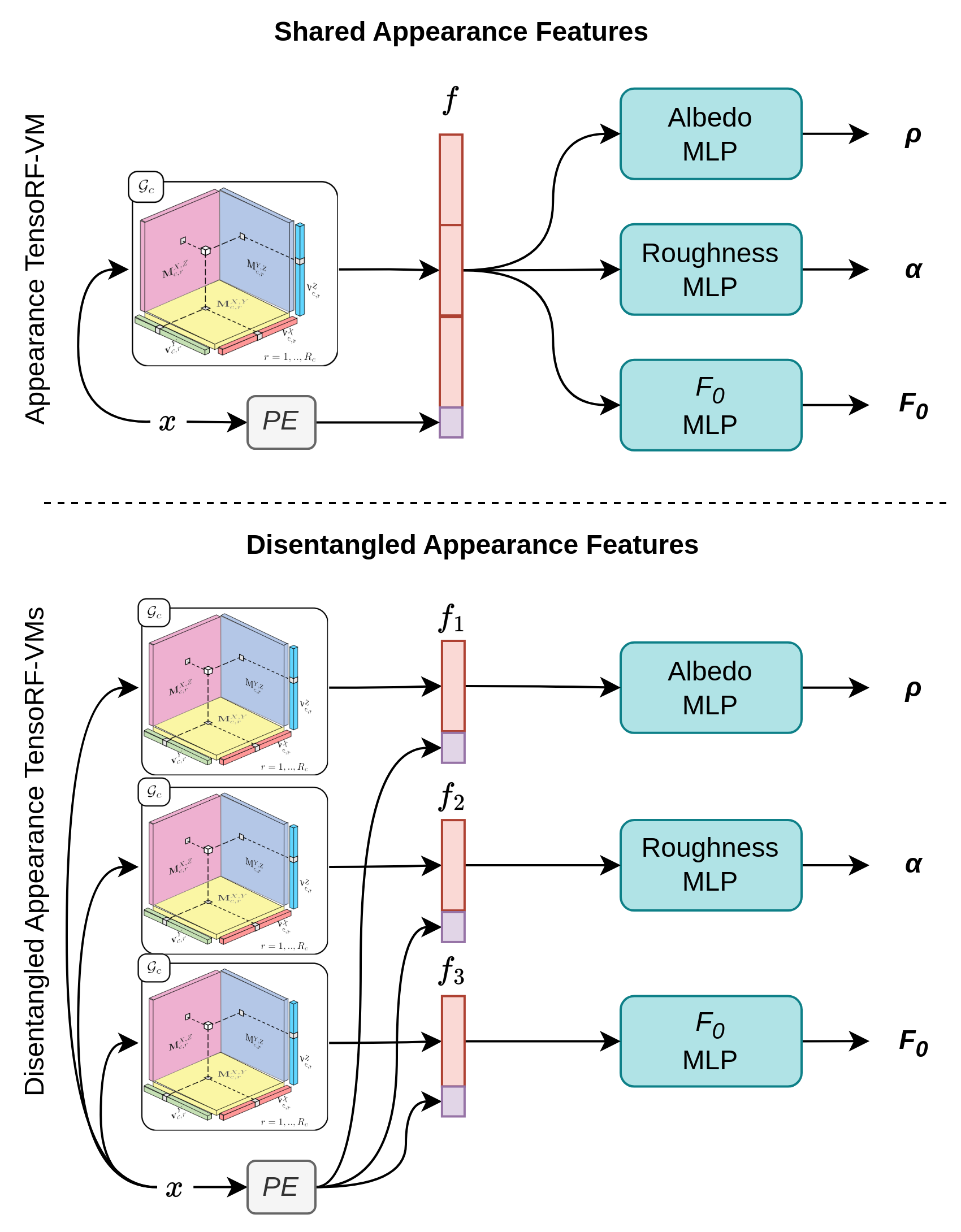}
    \caption{Disentanglement of appearance features to enable independent fine-tuning of albedo $\mathit{\rho}$, roughness $\mathit{a}$, and $\mathit{F_0}$. The single appearance TensoRF is split into three distinct TensoRFs with one third of the original feature dimensionality. This ensures that fine-tuning the albedo, roughness, or $F_0$ will not impact the input feature space of the other two.
    The TensoRF visualization was adapted from the original paper \cite{Chen2022tensorf}.}
    \label{fig:disentanglement}
\end{figure}

\noindent\textbf{Disentanglement.} 
As shown in the upper part of Fig.~\ref{fig:disentanglement}, albedo, roughness and $F_0$ MLPs are conditioned on the same appearance features extracted from a TensoRF representation $\mathcal{G}_c$. To facilitate independent fine-tuning of a single material property at a time, we first need to disentangle their inputs, considering that any gradients passing through an MLP would backpropagate to the shared appearance features and hence elicit change to the other properties as well. Simply splitting the feature vector into parts would still be insufficient, since the elements of the feature vector are not completely disentangled. Therefore, we replace the appearance TensoRF with three distinct TensoRFs but with proportionally smaller dimensions (from 24 to 8). The input feature vector of each MLP is also concatenated with the positional encoding (PE) of the 3D position $\boldsymbol{x}$. The adapted architecture is shown at the bottom of Fig.~\ref{fig:disentanglement}. \\

\noindent\textbf{Properties Tweaking.}  
To investigate the interaction between scene properties and discover any inherent compensating mechanisms, we first need to manipulate them by tweaking one property at a time before fine-tuning another. We employ three types of applied noise depending on the type of each property split into material, geometry, and illumination noise.
Inverse rendering models tend to underestimate or overestimate scene properties like albedo, roughness, and $F_0$ in NMF. To simulate such effects we apply a simple multiplier $m$ on the outputs of the MLPs, after the sigmoid, while also clipping the adjusted property so that it falls in the expected range. This entails the limitation that the model cannot fine-tune a property to reach a higher value than the multiplier and also if a property is closed to zero then a multiplier would have little effect. Multiplier values lower or greater than one simulate underestimation or overestimation of the property, respectively.
An additional common issue observed in inverse rendering, especially when dealing with shiny objects, is that an overestimation of roughness might result in a rougher more noisy geometry. To simulate this effect, we perturb the geometry by directly applying Gaussian noise on $\mathcal{G}_\sigma$, noted as $N(0,{\sigma}_d^2)$, where ${\sigma}_d$ stands for the variance of the distribution. On the other hand, underestimating the roughness may cause an estimation of a blurry environment map, which we can easily simulate by applying a Gaussian smoothing filter. We denote the filter as $G(s_I,\sigma_I)$, where  $s_I$ and $\sigma_I$ denote the size and the standard deviation of the Gaussian kernel, respectively. 


\section{Results}
\label{sec:results}

\subsection*{Evaluation Protocol}
We evaluate our experiments both qualitatively and quantitatively using conventional novel view synthesis metrics, namely PSNR \cite{10.5555/556230}, SSIM \cite{journals/tip/WangBSS04}, and LPIPS \cite{zhang2018unreasonable} for the visual aspect of the reconstruction. For the estimated geometry, we provide the mean angular error (MAE) in degrees between the estimated and ground truth surface normals. Last but not least, we also utilize PSNR to evaluate the quality of the estimated environment map compared to the ground truth as implemented in the released codebase of NMF\cite{mai2023nmf}. To differentiate between the PSNRs we term the PSNR of the environment map as EPSNR.

\subsection*{Consistency Across Illuminations}
\begin{figure*}[t]
    \centering
    \includegraphics[trim=15 0 10 0,clip,width=\textwidth]{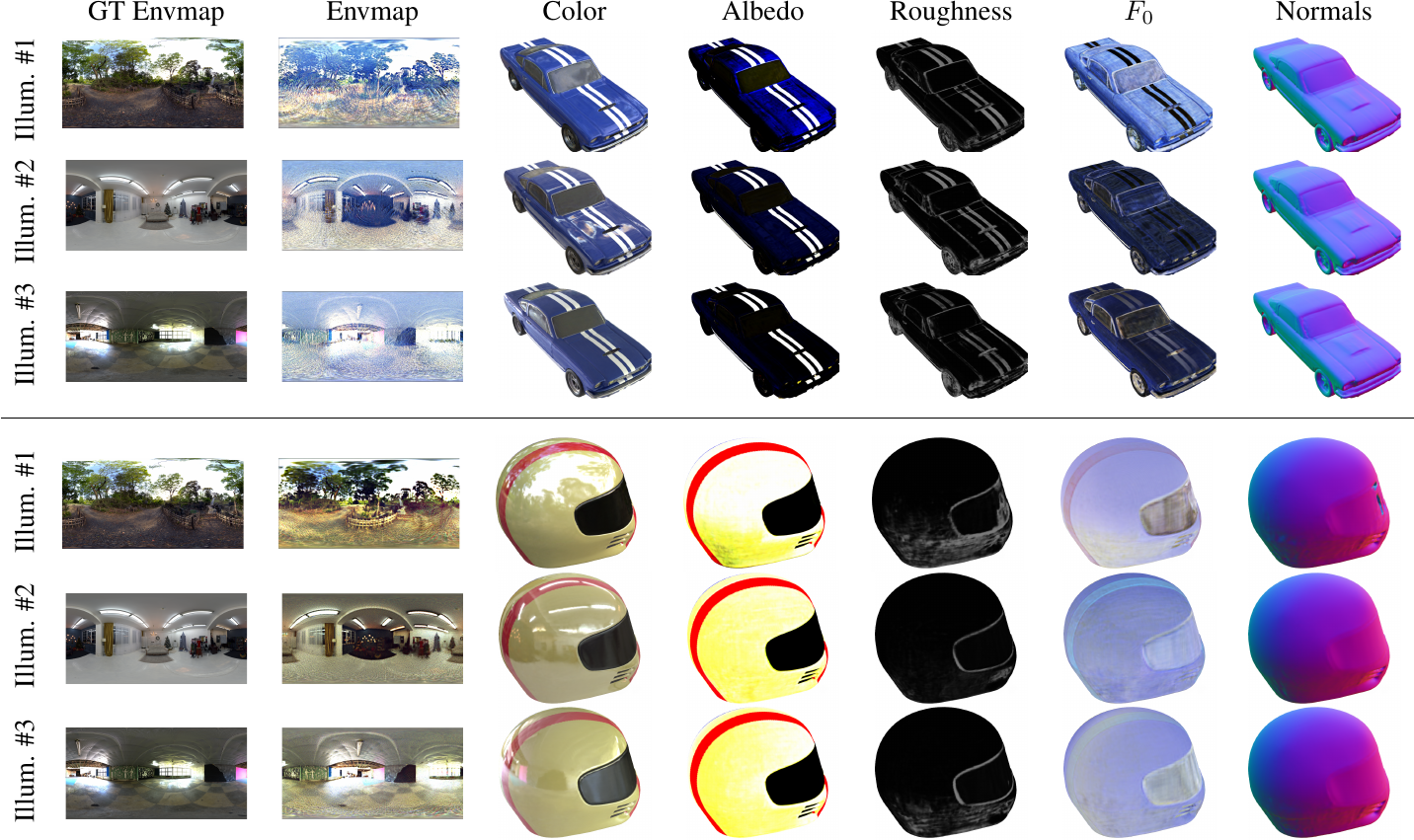}
    \caption{Estimated scene properties of the car and helmet scenes from the shiny-blender dataset under different illuminations. Due to the underconstrained nature of the problem and NMF optimizing solely for reconstruction, it fails to learn consistent scene properties.}
    \label{fig:inconsistencies}
\end{figure*}

In our first experiment, we evaluate the consistency of inverse rendering in estimating scene properties under different illuminations. To this end, we use two scenes from the shiny-blender dataset, namely \textit{car} and \textit{helmet}, and render each of them under three different illuminations using the released blender models \citep{verbin2022refnerf}. Then, we fully train NMF models for each object-illumination combination separately.
In \cref{fig:inconsistencies}, we present the estimated scene properties of each object under the different illumination settings as seen in the first column. The second column shows the estimated environment map, followed by the color renderings, albedo, roughness, $F_0$, and surface normals. As evident, NMF fails to estimate consistent scene properties across illuminations, with the largest variations occurring in albedo and $F_0$. The roughness and the geometry estimates (as viewed through surface normals) appear more consistent, although, there seem to be more errors in cases of high-frequency reflections as with \textit{Illumination \#1}. Furthermore, there seems to be bleeding of the albedo of the object into the estimate of the environment map, observed as the blue and yellow tints for the car and helmet, respectively. 
Based on the results of this experiment, we can safely conclude that inverse rendering under a single illumination, as in the case of NMF, can be an ill-posed problem with a large degree of inherent ambiguity in the estimation of the scene properties.


\subsection*{Scene Property Interactions and Compensations}
To evaluate the interaction between scene properties and their corresponding abilities to compensate for perturbations in other properties, we begin by fully training an NMF model for 30K iterations. Then, we manipulate one of the scene properties, freeze it, and successively unfreeze one other property at a time for fine-tuning. This way we can observe how effectively the latter can compensate for the perturbations introduced in the former.

To avoid the fine-tuning of one property affecting another, we completely disentangle albedo, roughness and $F_0$ as described in Section~\ref{sec:methodology}. In \cref{tab:shared_vs_disentangled}, we present the performance of NMF with shared versus disentangled appearance features across three scenes of the shiny blender dataset \cite{verbin2022refnerf}, namely ball, car, and helmet. Based on these results, we observe a minimal change in performance.



\begin{figure*}[t]
    \centering
    \centering
    \includegraphics[trim=40 40 0 0,clip,width=.2\linewidth]{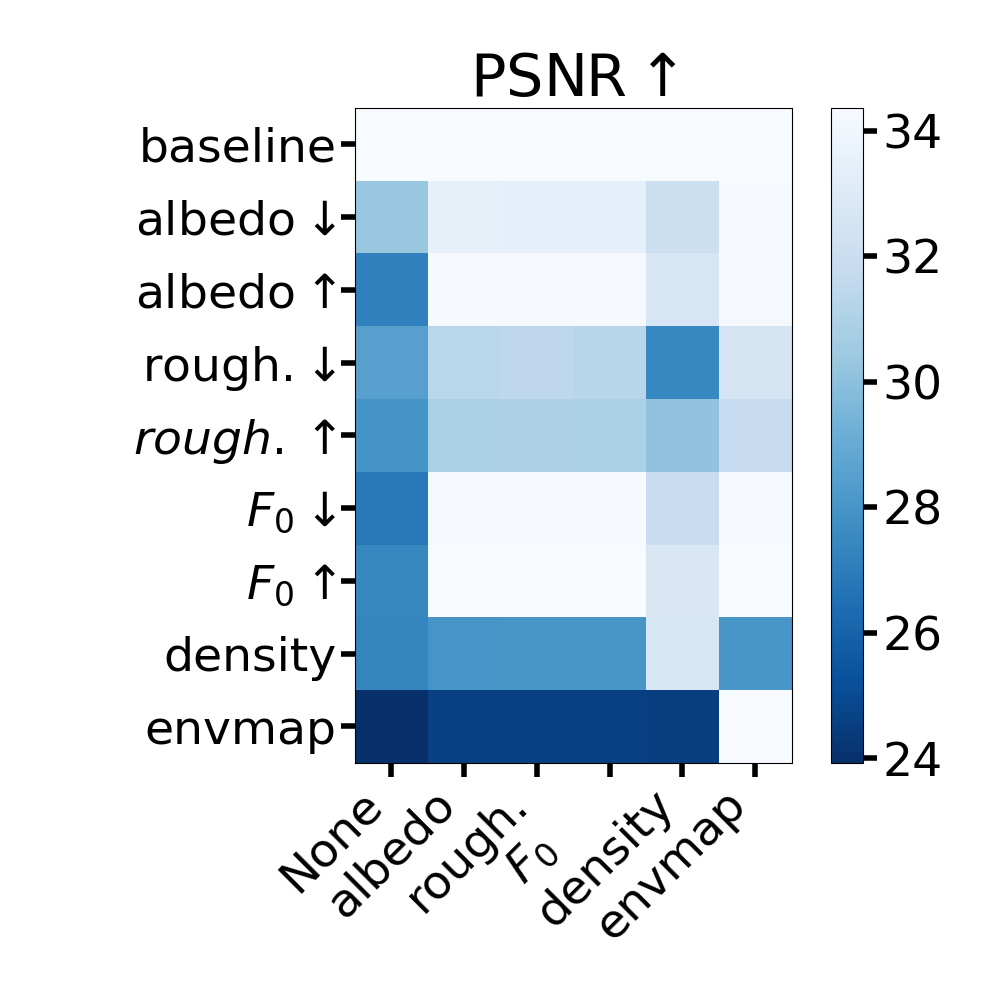}%
    \includegraphics[trim=40 40 0 0,clip,width=.2\linewidth]{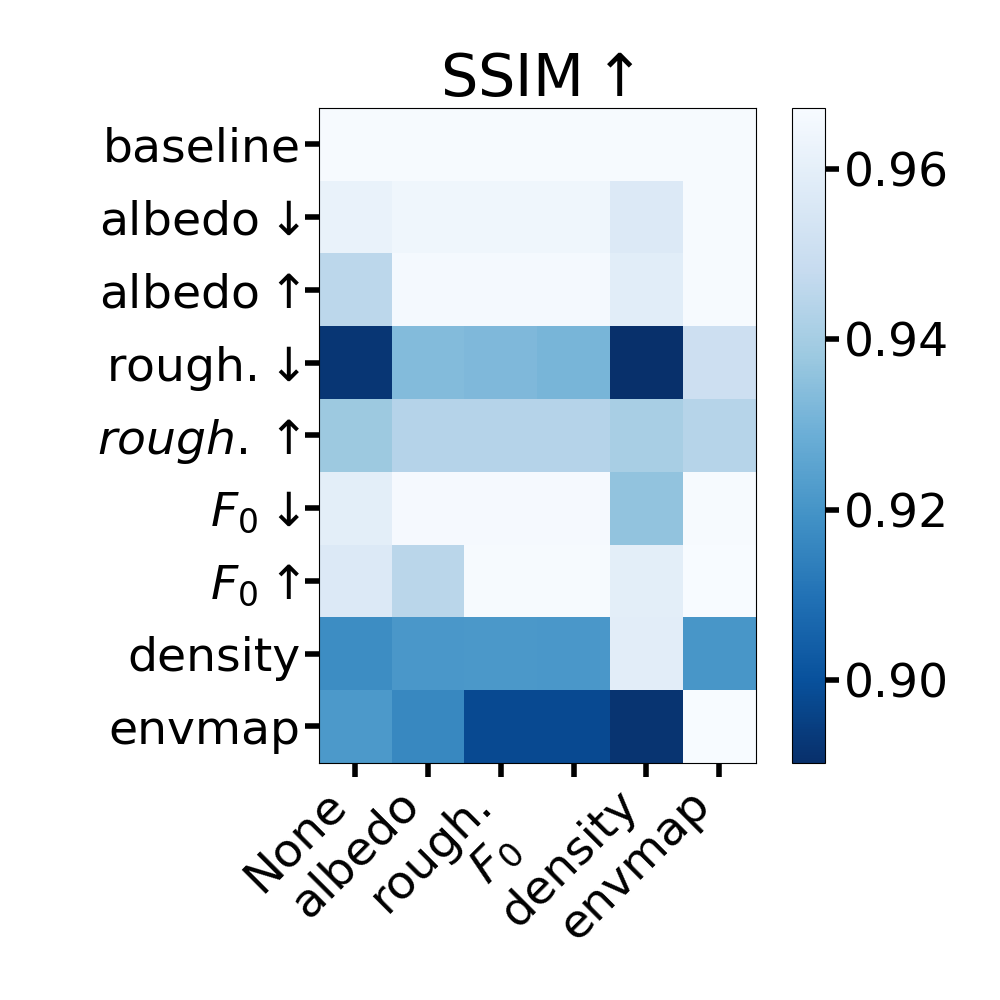}%
    \includegraphics[trim=40 40 0 0,clip,width=.2\linewidth]{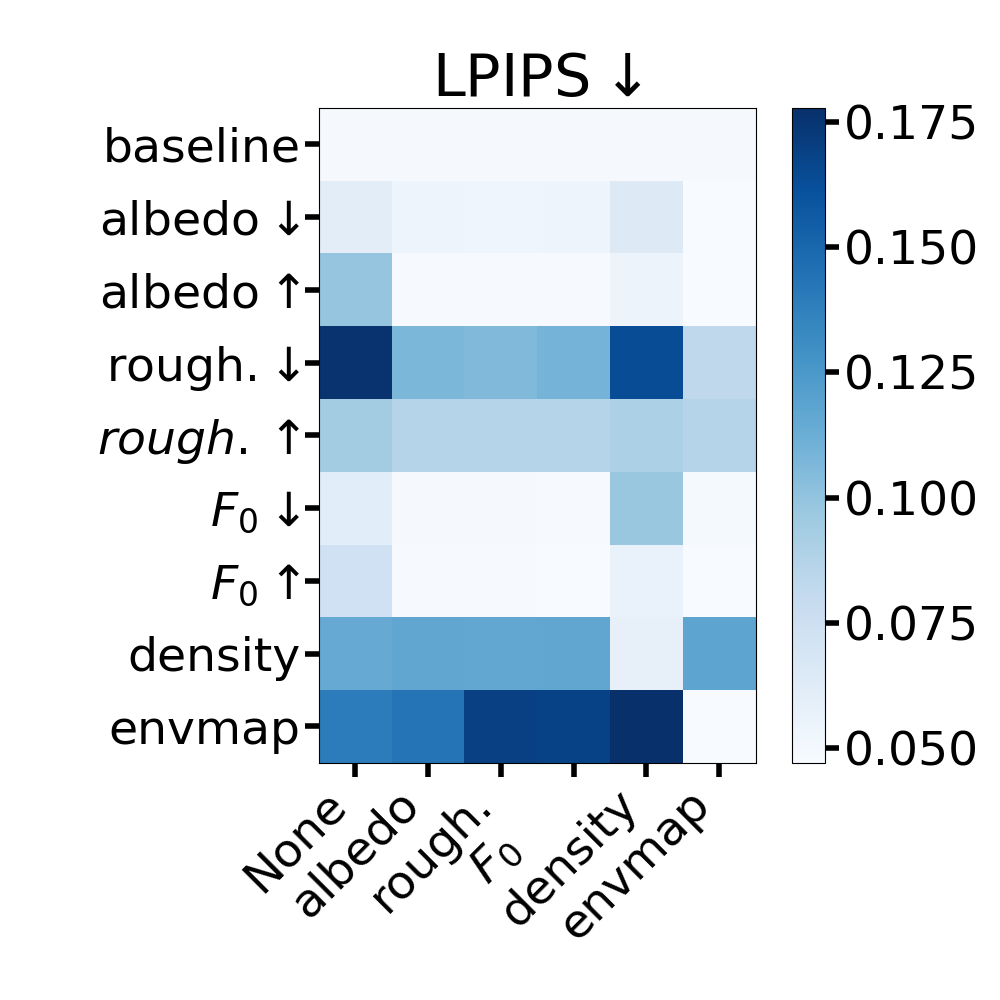}%
    \includegraphics[trim=40 40 0 0,clip,width=.2\linewidth]{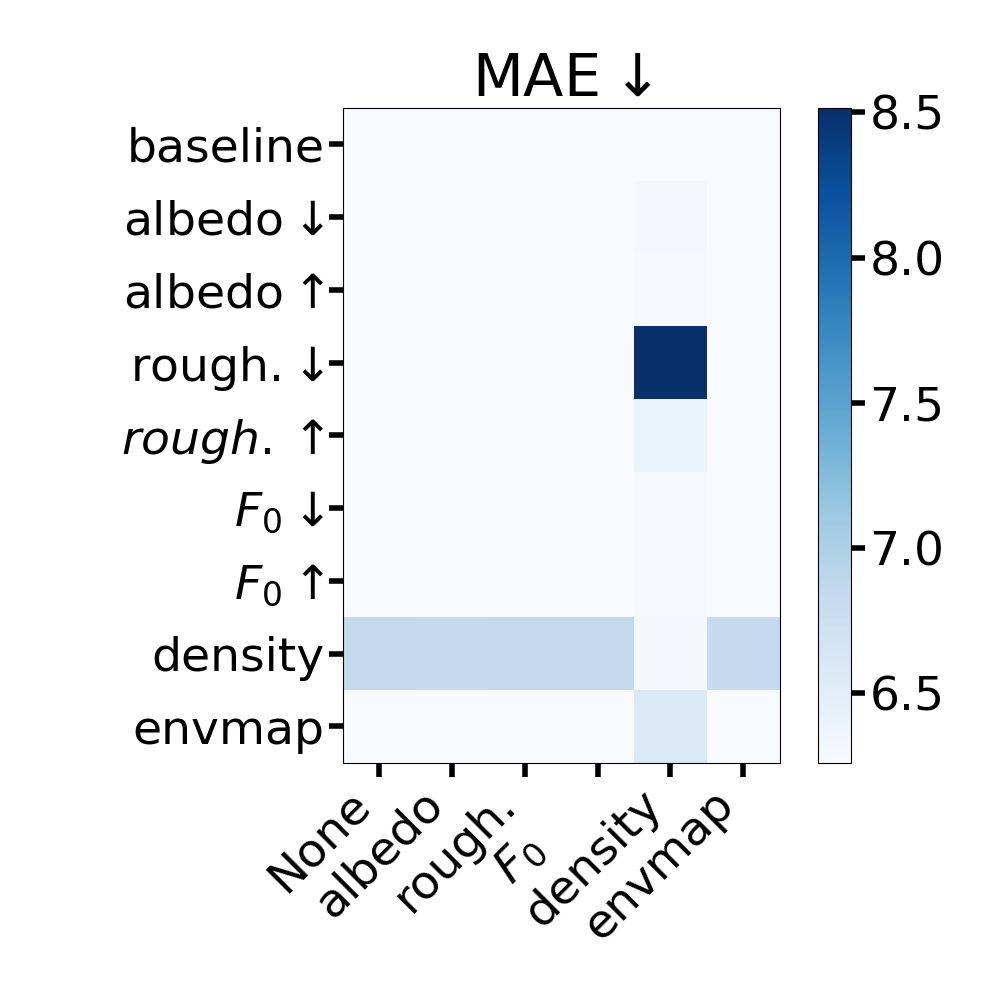}%
    \includegraphics[trim=40 40 0 0,clip,width=.2\linewidth]{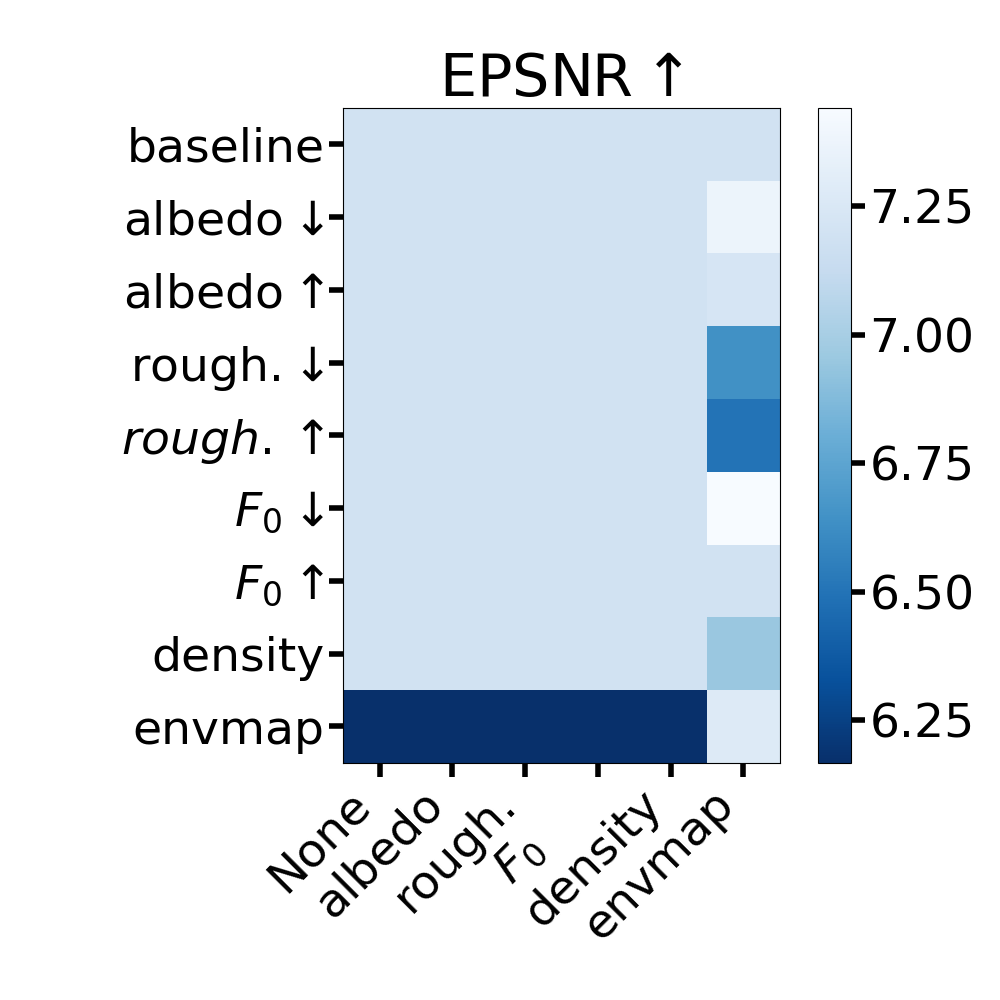}%
    \caption{Graphs demonstrating to what extent one scene property can compensate for suboptimal estimation of other properties. The vertical axis corresponds to the manipulated scene property while the horizontal axis corresponds to the fine-tuned property. The arrows $\uparrow, \downarrow$ on the manipulated property labels refer to overestimating or underestimating the property, respectively. A weighted average is used across the examined scenes (ball, car, helmet) of the shiny-blender dataset.
    }
    \label{fig:interactions}
\end{figure*}

\begin{figure*}[t]
    \centering
    \includegraphics[width=\textwidth]{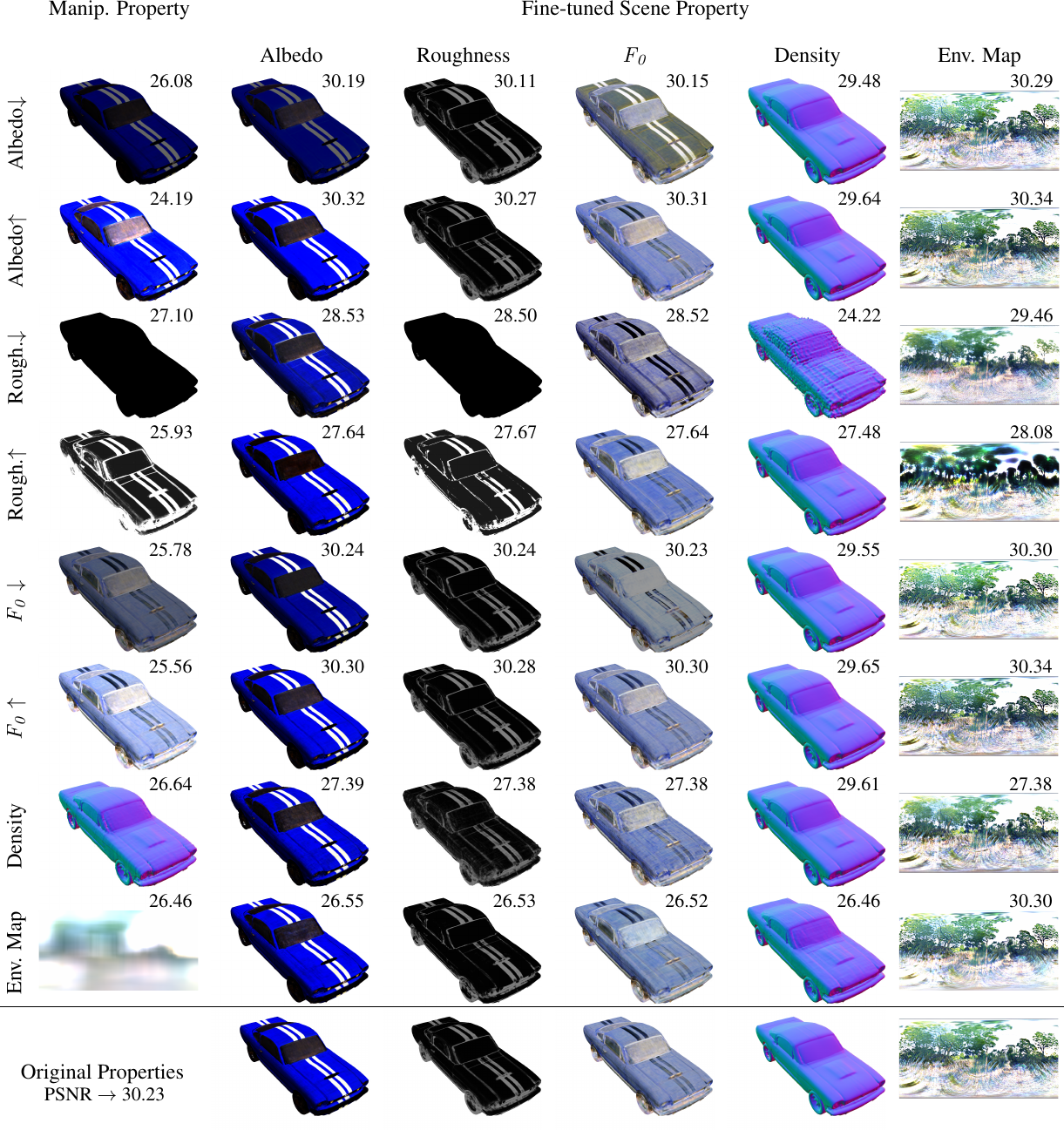}
    \caption{Effect of manipulating one scene property before fine-tuning another for the car scene of the shiny-blender dataset \cite{verbin2022refnerf}. The arrows $\downarrow, \uparrow$ next to albedo, roughness, and $F_0$ denote underestimation or overestimation respectively. The bottom row contains the original properties before adding noise and fine-tuning. For every experiment, we also include the PSNR at the top right corner of the corresponding image.}
    \label{fig:finetune-image-grid}
\end{figure*}

\cref{tab:noisy-metrics} shows the amount of manipulation applied on each scene property and the resulting performance drop before fine-tuning. The introduced noise to the model was tuned to reach a $15$-$25\%$ drop in PSNR. The results of the experiments are presented quantitatively for all scenes in \cref{fig:interactions} and qualitatively for just the car scene in \cref{fig:finetune-image-grid}. The quantitative results per scene and the qualitative results for the ball and helmet scenes are included in the supplementary material.

From the averaged quantitative results for all scenes in \cref{fig:interactions}, we observe that the model can easily compensate for changes in albedo and $F_0$, but struggles to improve performance when there are errors in roughness, density, or the environment map. Errors in density and the environment map specifically appear to be the hardest to recover from by compensating with another property. Moreover, it is evident across all scenes, that the model can better compensate for errors in albedo, roughness, and $F_0$ by fine-tuning one of them or the environment map but fine-tuning density, on the other hand, provides less improvement and results in worse geometry.  Additional results per scene are included in the supplementary material.

Also in \cref{fig:finetune-image-grid}, we can observe specific interactions and compensations between some scene properties. For example, underestimating albedo and fine-tuning $F_0$, seems to cause a drastic change in the latter, compared to the original unperturbed property. In general, fine-tuning $F_0$ appears to result in significantly more variation across the experiments compared to the other properties. Albedo and roughness mostly vary in intensity without any observable structural changes when they are fine-tuned. Fine-tuning the geometry, on the other hand, appears to vary most drastically when the roughness is manipulated or the environment map is smoothed. More specifically, a large degree of underestimation in roughness causes the geometry to become rougher itself, observed as the extremely noisy normals with correspondingly low metrics. Overestimation of the roughness, on the other hand, has less effect, although still noticeable, on the geometry. but an even larger effect on the environment map, causing it to become blurrier. At the same time, it seems that overestimating the smoothness of the environment map causes a rougher more noisy geometry as well. Lastly, we also note that the largest PSNR improvement is mostly observed when fine-tuning the environment map, which indicates its ability to compensate for perturbations on other properties.


\subsection*{Future Research}
Alleviating the ill-posedness of inverse rendering is certainly a difficult problem, especially for constant illumination scenes because of the complexity of disambiguating between geometry, materials, and illumination. However, priors can be beneficial to that end and therefore we highlight potential research directions focusing on guiding or regularizing the geometry, materials, or illumination properties of the scene. 

As observed in our experiments, suboptimal geometry leads to significant drops in performance that cannot be adequately compensated for by other scene properties. Therefore, accurate geometry estimation is vital to the inverse rendering process, and geometry priors can thus be beneficial in alleviating some ambiguity. 
Simple smoothing priors can easily be applied to the depth or normals, especially for shiny objects that can be assumed by definition to be locally smooth. However, this requires careful tuning to avoid oversmoothing and losing high-frequency geometry details. More sophisticated geometry priors would be either instance-specific (e.g. pretraining to extract a rough or even fine geometry as an initialization to speed up inverse rendering) or category-specific which would require training on a large amount of objects to generalize across categories and at the same time restrict the geometry search space. 

Regarding material priors and based on our experiments, we conclude that the suboptimal material properties can lead to large errors in geometry or illumination (especially for roughness) and even though the model may be able to compensate for such errors towards a better reconstruction by adapting albedo or $F_0$, subsequent relighting efforts would be heavily impacted. We also observed inconsistencies across surface points that should have the same material properties. As a result, we advocate for moving forward by utilizing pre-trained BRDFs \cite{liang2023envidr} to restrict the search space of inverse rendering to real physically plausible materials either explicitly or as a regularization step that penalizes unrealistic materials.
At the same time, another promising direction involves using priors to enforce a fixed number of materials thus restricting the search space and leading to more consistent materials across the surface of an object.

Finally, based on the outcomes of our experiments, the model struggles to estimate an accurate illumination due to the ambiguity with other scene properties which may cause high-frequency noise, oversmoothing, or bleeding of albedo into the environment map.
This indicates a need for illumination priors to improve this ambiguity. Specifically, we propose that such priors should follow the paradigm of \cite{lyu2023diffusionprior} and focus on generative approaches to guide the optimization by sampling more natural illuminations or more plausible illumination-material combinations. 

\begin{table}[t]
    \centering
    \setlength{\tabcolsep}{1pt}
    \caption{Performance of NMF with shared versus disentangled appearance features on three scenes of the shiny-blender dataset (ball, car, and helmet) and averaged over five runs. The table includes the average metric (black) and the standard deviation (blue) per scene and across the scenes. The small performance drop observed with the disentangled representation of the appearance features is deemed acceptable for the purposes of our experiments.}
    \label{tab:shared_vs_disentangled}
    \resizebox{\linewidth}{!}{\begin{tabular}{c c | c c c c}
    \hline
    & Parameter & PSNR & SSIM & LPIPS & MAE\\
    \hline\hline
    \multirow{3}{*}{\rotatebox[origin=c]{90}{shared}} 
     & ball & 38.73 \color{blue}{$\pm$ 0.042} & 0.982 \color{blue}{$\pm$ 0.000} & 0.051  \color{blue}{$\pm$ 0.000} & 2.492 \color{blue}{$\pm$ 0.004} \\
     & car & 30.04 \color{blue}{$\pm$ 0.070} & 0.949 \color{blue}{$\pm$ 0.001} & 0.033 \color{blue}{$\pm$ 0.000} & 7.709 \color{blue}{$\pm$ 0.085} \\
     & helmet & 34.18 \color{blue}{$\pm$ 0.100} & 0.970 \color{blue}{$\pm$ 0.000} & 0.052 \color{blue}{$\pm$ 0.000} & 2.415 \color{blue}{$\pm$ 0.028} \\
     \hline
     & Mean & 34.32 \color{blue}{$\pm$ 0.050} & 0.967 \color{blue}{$\pm$ 0.001} & 0.045 \color{blue}{$\pm$ 0.000} & 4.205 \color{blue}{$\pm$ 0.041}\\
     \hline
     \hline
     \multirow{3}{*}{\rotatebox[origin=c]{90}{disent.}}
     & ball & 38.46 \color{blue}{$\pm$ 0.032 }  & 0.976 \color{blue}{$\pm$ 0.013}  & 0.051 \color{blue}{$\pm$ 0.001}  & 2.482 \color{blue}{$\pm$ 0.004}\\
     & car & 30.20 \color{blue}{$\pm$ 0.024} & 0.951 \color{blue}{$\pm$ 0.000} & 0.03 \color{blue}{$\pm$ 0.000} & 8.552 \color{blue}{$\pm$ 0.025} \\
     & helmet & 33.56 \color{blue}{$\pm$ 0.166} & 0.964 \color{blue}{$\pm$ 0.004} & 0.056 \color{blue}{$\pm$ 0.002} & 2.626 \color{blue}{$\pm$ 0.061}\\
     \hline 
     & Mean & 34.07 \color{blue}{$\pm$ 0.070} & 0.964 \color{blue}{$\pm$ 0.006} & 0.046 \color{blue}{$\pm$ 0.001} & 4.553 \color{blue}{$\pm$ 0.030}\\
     \hline
    \end{tabular}}    
\end{table}

\begin{table}[t] 
\begin{threeparttable}[b]
    \centering
    \setlength{\tabcolsep}{1pt}
    \caption{Effect of manipulating a single parameter at a time without finetuning. The arrows $\downarrow, \uparrow$ next to albedo, roughness, and $F_0$ denote underestimation and overestimation, respectively. Albedo, roughness, and $F_0$ are perturbed using a multiplier $m$ at the output of the MLP after the sigmoid nonlinearity, while the geometry and illumination are perturbed by added Gaussian noise $N(0,\sigma_d^2)$ and Gaussian blurring $G(s_I, \sigma_I)$, respectively.}
    \label{tab:noisy-metrics}
    \begin{tabular}{c c | c | c c c c c}
    \hline
     & Parameter & Noise & PSNR & SSIM & LPIPS & MAE & EPSNR\\
    \hline\hline
    \multirow{9}{*}{\rotatebox[origin=c]{90}{ball}}
     & - & - &38.44 & 0.982 & 0.051 & 2.484 & 5.265 \\
     & Albedo$\downarrow^{\textcolor{red}{*}}$ & $m=0$ & 38.49 & 0.982 & 0.051 & 2.489 & 5.265 \\
     & Albedo$\uparrow$ & $m=1000$ & 30.53 & 0.959 & 0.127 & 2.484 & 5.265 \\
     & Rough.$\downarrow$ & $m=0.1$ & 29.40 & 0.892 & 0.236 & 2.484 & 5.265\\
     & Rough.$\uparrow$ & $m=2.0$ & 31.87 & 0.975 & 0.061 & 2.484 & 5.625 \\
     & $\mathit{F_0}\downarrow$ & $m=0.8$ & 28.69 & 0.979 & 0.055 & 2.484 & 5.265\\
     & $\mathit{F_0}\uparrow$ & $m=1.2$ & 30.79 & 0.980 & 0.054 & 2.484 & 5.265\\
     & Density & $N(0,1.05)$ & 29.05 & 0.936 & 0.128 & 3.142 & 5.265\\
     & Envmap & $G(301,300)$ & 19.88 & 0.904 & 0.214 & 2.484 & 5.099\\
    \hline
    \multirow{9}{*}{\rotatebox[origin=c]{90}{car}} 
     & - & - & 30.23 & 0.951 & 0.032 & 8.566 & 6.771 \\ 
     & Albedo$\downarrow$ & $m=0.4$ & 26.08 & 0.943 & 0.043 & 8.566 & 6.771\\
     & Albedo$\uparrow$ & $m=4.0$ & 24.19 & 0.933 & 0.046 & 7.566 & 6.771\\
     & Rough.$\downarrow$ & $m=0.0$ & 27.10 & 0.914 & 0.065 & 8.566 & 6.771 \\
     & Rough.$\uparrow$ & $m=10$ & 25.93 & 0.914 & 0.063 & 8.566 & 6.771\\
     & $\mathit{F_0}\downarrow$ & $m=0.6$ & 25.78 & 0.939 & 0.048 & 8.566& 6.771\\
     & $\mathit{F_0}\uparrow$ & $m=1.5$ & 25.56 & 0.939 & 0.055 & 8.566 & 6.771\\
     & Density & $N(0,1.5)$ & 26.64 & 0.911 & 0.060 & 8.970 & 6.771\\
     & Envmap & $G(151,100)$ & 26.46 & 0.922 & 0.060 & 8.566 & 6.189\\
     \hline
    \multirow{9}{*}{\rotatebox[origin=c]{90}{helmet}} 
     & - & - & 33.63 & 0.967 & 0.054 & 2.568 & 8.719\\ 
     & Albedo$\downarrow$ & $m=0.6$ & 24.95 & 0.96 & 0.077 & 2.568 & 8.719 \\
     & Albedo$\uparrow$ & $m=100$ & 26.12 & 0.945 & 0.099 & 2.568 & 8.719\\
     & Rough.$\downarrow^{\textcolor{red}{*}}$ & $m=0$ & 28.86 & 0.871 & 0.174 & 2.568 & 8.719\\
     & Rough.$\uparrow$ & $m=15$ & 25.34 & 0.925 & 0.138 & 2.568 & 8.719\\
     & $\mathit{F_0}\downarrow$ & $m=0.6$ & 25.82 & 0.960 & 0.073 & 2.568 & 8.719\\
     & $\mathit{F_0}\uparrow$ & $m=1.5$ & 25.25 & 0.951 & 0.099 & 2.568 & 8.719\\
     & Density & $N(0,3)$ & 26.18 & 0.906 & 0.128 & 3.649 & 8.719\\
     & Envmap & $G(151,100)$ & 26.22 & 0.939 & 0.110 & 2.568 & 6.809 \\
    \hline
    \end{tabular}
    \begin{tablenotes}
       \item[\textcolor{red}{*}] Properties are too low 
       for an underestimation experiment to have the desired drop in performance before finetuning, but we still report them for completeness.
    \end{tablenotes}
\end{threeparttable}
\end{table}

\section*{Conclusion}
\label{sec:conclusion}
In this work, we investigate the ambiguity and ill-posedness of neural inverse rendering while trying to uncover interactions between the scene properties and potential compensating mechanisms that allow a model to produce plausible reconstructions with suboptimal geometry, material properties, and illumination.
Using Neural Microfacet Fields (NMF), we showcase this ambiguity by demonstrating how the scene properties can vary under different lighting conditions. 
Given our proposed evaluation framework, we perform controlled experiments for manipulating one scene property and independently fine-tuning another to discover interactions and compensations between them.
Our experiments demonstrate that the model can easily compensate for suboptimal albedo or $F_0$ estimates, but not so easily for roughness, geometry, or illumination which can significantly negatively impact results.
Finally, we propose future research towards geometry, material, and illumination priors to reduce the ambiguity of inverse rendering tasks and produce more consistent results across different illuminations, and thus more accurate, plausible, and consistent relighting.

\section*{Acknowledgements}
We gratefully acknowledge funding support from the Sim2Real2 project, in the context of the Ford-KU Leuven alliance program.

\clearpage\clearpage
{
\small
\bibliographystyle{ieeenat_fullname}
\bibliography{main}
}

\end{document}